# Polyp-DDPM: Diffusion-Based Semantic Polyp Synthesis for Enhanced Segmentation

Zolnamar Dorjsembe*[†], Hsing-Kuo Pao* and Furen Xiao[†]
Email: d11115806@mail.ntust.edu.tw, r12458015@ntu.edu.tw, pao@mail.ntust.edu.tw, fxiao@ntu.edu.tw
*Department of Computer Science and Information Engineering, National Taiwan University of Science and Technology, Taipei, Taiwan
[†]Institute of Medical Device and Imaging, National Taiwan University College of Medicine, Taipei, Taiwan

*Abstract*—This study introduces Polyp-DDPM, a diffusion-based method for generating realistic images of polyps conditioned on masks, aimed at enhancing the segmentation of gastrointestinal (GI) tract polyps. Our approach addresses the challenges of data limitations, high annotation costs, and privacy concerns associated with medical images. By conditioning the diffusion model on segmentation masks—binary masks that represent abnormal areas—Polyp-DDPM outperforms state-of-the-art methods in terms of image quality (achieving a Fréchet Inception Distance (FID) score of 78.47, compared to scores above 83.79) and segmentation performance (achieving an Intersection over Union (IoU) of 0.7156, versus less than 0.6694 for synthetic images from baseline models and 0.7067 for real data). Our method generates a high-quality, diverse synthetic dataset for training, thereby enhancing polyp segmentation models to be comparable with real images and offering greater data augmentation capabilities to improve segmentation models. The source code and pretrained weights for Polyp-DDPM are made publicly available at https://github.com/mobaidoctor/polyp-ddpm.

*Keywords*—diffusion models, semantic polyp synthesis, polyp segmentation

## I. INTRODUCTION

Colorectal cancer (CRC) ranks as the third most common and second deadliest cancer globally [1]. CRC often begins as colorectal polyps, which are early indicators of the disease. Detecting and removing these polyps early through colonoscopy can prevent CRC and lower death rates. However, identifying small polyps during colonoscopy can be difficult, depending on the expertise of doctors and other challenges, such as polyps being out of view or overlooked during the procedure [2].

To enhance polyp detection, researchers are employing machine learning to autonomously identify and emphasize polyps in endoscopies [3]. However, the development of these technologies faces significant challenges due to the necessity for extensive and varied datasets that are essential for training models to achieve high accuracy. The healthcare industry often faces a shortage of such data, attributable to the diversity in the appearance of abnormal areas, difficulties in recruiting patients, the high cost of data annotation, and concerns about patient data privacy [4].

To mitigate the data scarcity issue, the exploration of synthetic images as a viable solution has gained attention [5]. Thambawita et al. [6] developed a GAN-based method for creating polyp images with segmentation masks, using a two-stage process involving initial training on the HyperKvasir dataset [7] of 1,000 images, followed by style transfer to generate synthetic images. Despite achieving more realistic images than other GAN models, their SinGAN-Seg model faces challenges in diversity and detail accuracy. A prevalent issue with GAN models is the mode collapse problem.

Recent progress in diffusion-based models has overcome the mode collapse issue, producing diverse, high-quality images that outperform GANs [8]. Macháček et al. [9] introduced a two-stage diffusion model for polyp image and mask generation, using the Kvasir-SEG dataset [10]. This process involves generating masks with an improved diffusion model, then conditioning a latent diffusion model on these masks to create images. Despite its effectiveness in generating varied images, this method incurs high computational costs for training and inference due to the need for two models.

In response to these existing challenges, we introduce a novel diffusion-based semantic polyp synthesis method, Polyp-DDPM, aimed at enhancing polyp segmentation builds upon our previous work, Med-DDPM [11]. This approach conditions the diffusion model through channel-wise concatenation of mask images. We conducted experiments using the Kvasir-SEG dataset and compared our proposed method with SinGAN-Seg [6] and latent diffusion model [9], as these represent the latest advancements in annotated polyp dataset generation, encompassing both GAN-based and diffusion-based methods.

In our experiments, Polyp-DDPM demonstrated superior performance compared to baseline models in both image quality and segmentation tasks. This research contributes to the field by offering a new diffusion-based approach for synthesizing high-quality synthetic polyp images for any given mask images, which can be used to train more accurate polyp segmentation models. The source code and pretrained models are made publicly available to further research and application in this important area of medical imaging.

## II. METHODS

In this study, we build upon our previous work on semantic 3D brain MRI synthesis [11] and enhance the architecture to generate conditional 2D polyp images based on segmentation masks.

Our method involves the forward diffusion process $q$, small amounts of Gaussian noise $\epsilon \sim \mathcal{N}(0, I)$ defined by variance schedule $\bar{\alpha}_t$ are added to an image sample $x_0$ of training dataset at each timestep $t$ in a given number of timesteps T:

$$x_t = \sqrt{\bar{\alpha}_t} x_0 + \sqrt{1 - \bar{\alpha}_t} \epsilon. \quad (1)$$

To avoid sudden fluctuations in the noise level, a cosine noise schedule presented in [12] is adapted, defined as follows:

$$\bar{\alpha}_t = \frac{f(t)}{f(0)}, \qquad f(t) = \cos\left(\frac{t/T + s}{1 + s} \cdot \frac{\pi}{2}\right)^2, \quad (2)$$

where parameter $s$ represents a small offset value that prevents the schedule from being too small when the timestep gets closer to zero. [11]

In the reverse diffusion process $p_\theta$, we employ the U-Net architecture with 64 input channels as our denoiser model. The core architecture of our proposed method is depicted in Fig. 1. The main components of the denoiser U-Net architecture include sinusoidal position embeddings, which are utilized to encode the timestep $t$, thereby informing the model about the specific noise level affecting the input images. A key element of this architecture is the wide ResNet blocks, which consists of convolutional layers, fully connected layers, group normalization, SiLU activation layers, and skip connections. Group normalization integrates the 2D convolutional layers following the attention layers. To enable conditional modeling, we introduce a straightforward yet efficient technique that modifies the input image $x_t$ by concatenating the segmentation mask $c$ in a channel-wise manner.

## III. EXPERIMENTS AND RESULTS

We utilized the Kvasir-SEG dataset [10] with the same train and test splits as those used by LDM [9] to train our proposed method. The images were resized to 256x256 pixels, and the pixel intensity was rescaled to the range [-1, 1]. Our model was trained using 900 images and then tested on 100 test images. To ensure a fair comparison, we employed pretrained models of LDM [9] and SinGAN-Seg [6]. However, the SinGAN-Seg model, which was trained on 1000 images from the HyperKvasir dataset and incorporates style transfer, presents an unfair comparison with models trained on only 900 images without style transfer. Despite this, we aimed to assess the efficacy of our diffusion model against it. Our model was trained using 100,000 iterations with a learning rate of $10^{-4}$, a batch size of 32, 64 input channels, employing only 250 timesteps, and utilizing an L1 loss function. During training, we applied augmentation techniques such as rotation, horizontal flipping, and random rotation.

For polyp segmentation tasks, as a qualitative evaluation of synthetic images, we utilized the same segmentation models — UNet++, FPN, and DeepLabv3plus—as [9], with the same hyperparameter configurations except for altering the number of epochs to 100 and modifying the number of training and test images. All segmentation models were trained on 900 images from the Kvasir-SEG training set and 900 synthetic images created from the mask images of the training set. To evaluate the effectiveness of the synthetic images, we tested the segmentation models on 1000 images from the HyperKvasir dataset [7] and 196 images from the ETIS-LaribPolypDB dataset [13], along with 100 test images from the Kvasir-SEG dataset.

Synthetic images are quantitatively evaluated using the Fréchet Inception Distance (FID), Inception Score (IS), and Kernel Inception Distance (KID) scores by comparing 1,000 samples of both synthetic and real images. We conducted quantitative evaluations against two different datasets: Kvasir-SEG and HyperKvasir. The performance of segmentation models was assessed using the Intersection over Union (IoU), F1 Score, Accuracy, and Precision scores. All models were trained on a Tesla V100-SXM2 32 GB GPU card.

### A. Results of Image Synthesis

Fig. 2 presents a comparison between real and synthetic images generated by our proposed method and two baseline models, using a given mask from the HyperKvasir dataset. This dataset is utilized for training the SinGAN-Seg model and as unseen data for our proposed method and the Latent Diffusion Model (LDM). The comparison clearly demonstrates that the two diffusion models possess a higher capability to generate diverse samples compared to the SinGAN-Seg model. Although the GAN-Seg model, which uses input masks from its training data, should theoretically generate better images than the other two diffusion models, it is evident that the pretrained SinGAN-Seg model suffers from mode collapse and produces only slightly varied images. In contrast, the two diffusion models are capable of generating diverse, high-quality samples. When comparing our proposed Polyp-DDPM with the LDM, our model is able to generate more diverse images with precise details than the LDM.

In terms of quantitative evaluation, our proposed method, Polyp-DDPM, outperformed the other baseline models on both the Kvasir-SEG and HyperKvasir datasets (Table I).

TABLE I. QUANTATIVE RESULTS

| Method | Kvasir-SEG [10] | | | HyperKvasir [7] | | |
|---|---|---|---|---|---|---|
| | FID↓ | IS↑ | KID↓ | FID↓ | IS↑ | KID↓ |
| LDM [9] | 95.8243 | 2.3096 | 0.0920 | 97.0121 | 2.3096 | 0.0942 |
| SinGAN-Seg [6] | 141.1729 | **3.5943** | 0.1553 | 131.3347 | **3.3607** | 0.1468 |
| **Ours** | **78.4797** | 2.7361 | **0.0704** | **81.1045** | 2.7361 | **0.0755** |
| [10] vs [7] | 3.6111 | 3.2925 | 0.0000 | 3.6111 | 3.2925 | 0.0000 |

Our method achieved the lowest Fréchet Inception Distance (FID) score of 78.47 and Kernel Inception Distance (KID) score of 0.07 when compared with images from the Kvasir-SEG dataset, and an FID of 81.10 and KID of 0.07 on real images from the HyperKvasir dataset. In comparison, the LDM had the second best scores with an FID of 95.82 and

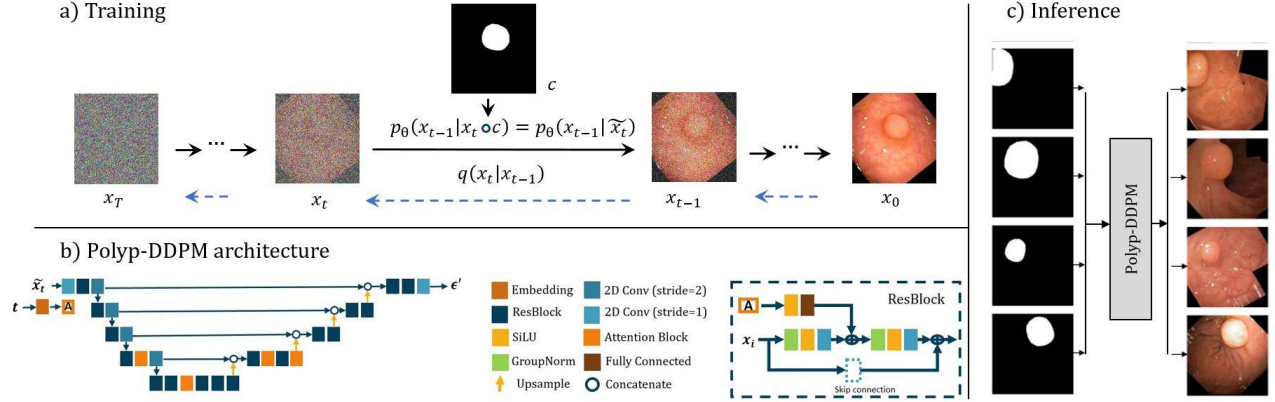

Fig. 1. Overall architecture of Polyp-DDPM for training and generating synthetic polyp images. a) Training: Polyp-DDPM is trained to transform random noise into realistic polyp images by conditioning on a binary segmentation mask of the abnormal area. b) Core building blocks of the Polyp-DDPM model. c) Inference: The trained Polyp-DDPM model performs inference on a given input mask to generate corresponding synthetic images.

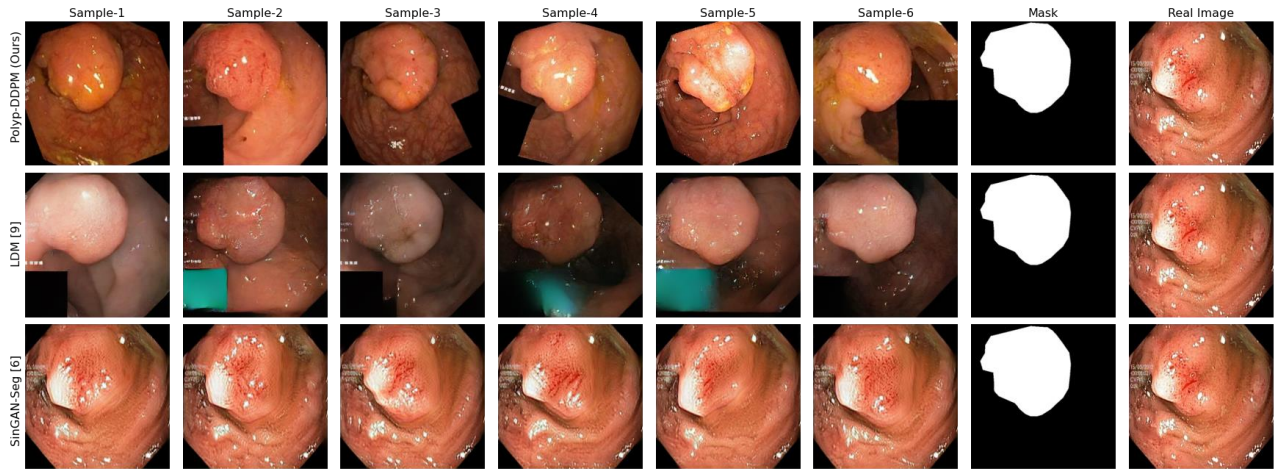

Fig. 2. Comparison of real and synthetic samples: showcasing the diversity of synthetic images generated from a single input mask

TABLE II. QUALITATIVE RESULTS

| # | Data | Model | UNet++ (26.1M) | | | | FPN (23.2M) | | | | DeepLabv3plus (22.4M) | | | |
|---|---|---|---|---|---|---|---|---|---|---|---|---|---|---|
| | | | IoU | F1 | Acc | Prec | IoU | F1 | Acc | Prec | IoU | F1 | Acc | Prec |
| Test set-1[a] | R900 | - | 0.7067 | 0.8281 | 0.9460 | 0.8383 | 0.7730 | 0.8720 | 0.9596 | 0.8786 | 0.7217 | 0.8384 | 0.9461 | 0.8012 |
| | S900 | LDM [9] | 0.6694 | 0.8020 | 0.9361 | 0.7905 | 0.6391 | 0.7798 | 0.9215 | 0.7035 | 0.6780 | 0.8081 | 0.9365 | 0.7779 |
| | | SinGAN-Seg [6] | 0.6828 | 0.8115 | 0.9372 | 0.7758 | 0.6560 | 0.7922 | 0.9386 | **0.8564** | 0.6520 | 0.7894 | 0.9343 | 0.8042 |
| | | **Polyp-DDPM (Ours)** | **0.7156** | **0.8342** | **0.9464** | **0.8203** | **0.7027** | **0.8254** | **0.9432** | 0.8075 | **0.6999** | **0.8235** | **0.9445** | **0.8324** |
| Test set-2[b] | R900 | - | 0.8921 | 0.9430 | 0.9825 | 0.9442 | 0.9105 | 0.9532 | 0.9856 | 0.9547 | 0.8903 | 0.9420 | 0.9818 | 0.9254 |
| | S900 | LDM [9] | 0.7167 | 0.8349 | 0.9493 | 0.8371 | 0.6743 | 0.8055 | 0.9330 | 0.7278 | 0.7162 | 0.8346 | 0.9475 | 0.8104 |
| | | SinGAN-Seg [6] | 0.7201 | 0.8373 | 0.9476 | 0.8024 | 0.6866 | 0.8142 | 0.9463 | **0.8704** | 0.6889 | 0.8158 | 0.9429 | 0.8095 |
| | | **Polyp-DDPM (Ours)** | **0.7739** | **0.8725** | **0.9604** | **0.8649** | **0.7735** | **0.8723** | **0.9600** | 0.8585 | **0.7723** | **0.8715** | **0.9603** | **0.8687** |
| Test set-3[c] | R900 | - | 0.6329 | 0.7752 | 0.9784 | 0.7343 | 0.6242 | 0.7686 | 0.9783 | 0.7425 | 0.5262 | 0.6895 | 0.9664 | 0.5926 |
| | S900 | LDM [9] | 0.3664 | 0.5363 | 0.9431 | 0.4250 | 0.2544 | 0.4056 | 0.8845 | 0.2644 | 0.2434 | 0.3915 | 0.8800 | 0.2541 |
| | | SinGAN-Seg [6] | 0.3209 | 0.4859 | 0.9457 | 0.4250 | 0.3150 | 0.4791 | 0.9445 | 0.4168 | 0.2358 | 0.3816 | 0.9022 | 0.2673 |
| | | **Polyp-DDPM (Ours)** | **0.5295** | **0.6923** | **0.9680** | **0.6131** | **0.5530** | **0.7122** | **0.9705** | **0.6373** | **0.5682** | **0.7246** | **0.9739** | **0.6940** |

[a.] Test 100 images of Kvasir-SEG dataset, [b.] 1000 images of HyperKvasir dataset, [c.] 196 images of ETIS-LaribPolypDB dataset.

KID of 0.09 on Kvasir-SEG, and an FID of 97.01 and KID of 0.09 on HyperKvasir.

Conversely, SinGAN-Seg recorded significantly higher FID and KID scores of 131.13 and 0.14, respectively, when compared to its own training dataset, HyperKvasir, and also performed poorly on the Kvasir-SEG dataset. However, SinGAN-Seg achieved the highest Inception Score compared to the two diffusion-based models, attributed to the ability of SinGAN-Seg model for style transfer from real images.

*B. Results of Segmentation Experiments*

We conducted a comprehensive analysis of the performance of three segmentation models—UNet++, FPN, and DeepLabv3plus—across three different test datasets: Kvasir-SEG, HyperKvasir, and ETIS-LaribPolypDB, comparing the efficacy of training with synthetic versus real images as detailed in Table II. When trained on 900 synthetic images, the Polyp-DDPM model exhibited remarkable results on the test set of the Kvasir-SEG Dataset. Specifically, with the UNet++ model, it achieved an IoU of 0.7156, an F1 score of 0.8342, an accuracy of 0.9464, and a precision of 0.8203, surpassing both the SinGAN-Seg and LDM models. This improvement was even more pronounced compared to when trained with 900 real images, where the IoU was 0.7067 and the F1 score was 0.8281, highlighting the efficacy of synthetic data in enhancing segmentation performance. A similar trend was observed with the FPN and DeepLabv3plus models, where Polyp-DDPM outperformed the other two baseline models. However, the IoU and F1 scores for the FPN and DeepLabv3plus models were lower than those compared to the real image results.

On the unseen HyperKvasir dataset, the superiority of Polyp-DDPM was even more evident. With the UNet++ model, it achieved an IoU of 0.7739 and an F1 score of 0.8725, surpassing the performance of the SinGAN-Seg model trained on the entire HyperKvasir dataset and also outperforming the LDM. This trend continued with the FPN and DeepLabv3plus models, and even further with the unseen ETIS-LaribPolypDB dataset, emphasizing the ability of Polyp-DDPM to better generalize. SinGAN-Seg images on both the Kvasir-SEG and HyperKvasir datasets, the only precision scores of the FPN models were higher than those of other models. However, this pattern was not observed on the ETIS-LaribPolypDB dataset.

Nevertheless, when applied to these unseen datasets, the performance of synthetic images did not match that of real images, leading to lower scores. This highlights the need for further improvements in the quality of synthetic images. In addition, we investigated the use of a combination of real and synthetic images for training segmentation models and found considerable promise in the data augmentation capabilities of our proposed method. For instance, a mixed training set of 1800 images (900 real and 900 Polyp-DDPM) achieved an IoU of 0.7484 and F1 score of 0.8561. In contrast, using only 900 real images, the IoU and F1 scores were lower: 0.7067 and 0.8281 for UNet++, respectively. Similarly, for DeepLabv3plus, the mixed set yielded an IoU of 0.7496 and F1 of 0.8569, surpassing the IoU of 0.7217 and F1 of 0.8384 with real images. However, due to page limitations, we have excluded these experimental results from this paper and made them available in our GitHub repository (https://github.com/mobaidoctor/polyp-ddpm).

IV. CONCLUSION

This study introduces Polyp-DDPM, a novel diffusion-based semantic polyp synthesis method that has been shown to outperform existing GAN-based and diffusion-based models in generating high-quality, diverse synthetic polyp images. The quantitative assessments, using the Fréchet Inception Distance and Kernel Inception Distance metrics, further substantiate the superiority of Polyp-DDPM over the SinGAN-Seg and Latent Diffusion Model, especially in generating images that closely resemble real dataset characteristics. Furthermore, the segmentation experiments underscore the potential of synthetic images generated by our proposed method to improve the training of polyp segmentation models, making them comparable to real images and achieving notable results across various test datasets. The comparative advantage of Polyp-DDPM is particularly pronounced in its ability to generate images with greater diversity and precision, thereby addressing the critical challenge of data scarcity in the medical imaging field. This research not only advances the technological frontier in synthetic image generation but also paves the way for more effective and accessible medical imaging solutions, ultimately contributing to improving the training of models for the early detection and prevention of colorectal cancer.